# CREER: A Large-Scale Corpus for Relation Extraction and Entity Recognition

**Yu-Siou Tang, Chung-Hsien Wu**
National Cheng Kung University, Tainan, Taiwan
p76084481@ncku.edu.tw, chunghsienwu@gmail.com

## Abstract

We describe the design and use of the CREER dataset, a large corpus annotated with rich English grammar and semantic attributes. The CREER dataset uses the Stanford CoreNLP Annotator to capture rich language structures from Wikipedia plain text. This dataset follows widely used linguistic and semantic annotations so that it can be used for not only most natural language processing tasks but also scaling the dataset. This large supervised dataset can serve as the basis for improving the performance of NLP tasks in the future. We publicize the dataset through the link: https://140.116.82.111/share.cgi?ssid=000dOJ4

## 1 Introduction

This paper describes the design and motivation of the CREER dataset, which contains aspects of human language, including syntax (such as parse trees and dependency parsing) and world knowledge information (such as entity mentions and relations). We describe the design of the dataset and its original motivation (Section 2), how the data was collected (Section 3), the set of attributes in the dataset (Section 4), and conclusions (Section 5). While there are several popular tasks- or domain-specific datasets, the CREER dataset covers a large number of natural language attributes in large-scale text. The dataset is available through the link: https://140.116.82.111/share.cgi?ssid=000dOJ4

Table 1 shows the statistics of CREER and existing datasets. Most corpora with knowledge annotations are used for downstream tasks such as named entity recognition and relation extraction. CREER dataset scales up a corpus with knowledge annotation, leading to usage for pre-training. Thus, the CREER dataset has a much larger number of labels on plain text compared with existing datasets containing both entity and relation labels. CREER dataset has four advantages:

1. This large dataset can benefit from supervised learning methods.
2. The properties in the CREER dataset are widely used, so it is easy to extend with existing datasets.
3. The dataset reflects the complexity and volatility of real-world text since the sentence length in the CREER dataset is diverse.
4. The CREER dataset provides a large-scale knowledge-centric dataset, and it can be used as a good testbed to facilitate the development of these basic NLP tasks.

## 2 Design and Original Motivation

The CREER dataset is designed for training word representation with knowledge information. Numerous natural language processing (NLP) tasks aim to analyze human language, including syntax (e.g., constituency and dependency parsing), semantics (e.g., semantic role labeling), factual knowledge (e.g., named entity recognition and relation extraction) or sentiment analysis. While learning contextual word representation (Devlin et al., 2018;Matthew E. Peters, 2018) can use large plain text in an unsupervised manner, previous works have shown that learning word representation using various aspects of human language can effectively improve the performance of many NLP tasks. However, previous studies have had little exploration of the way that learn factual knowledge from context, and this is the gap that we aim to explore. Thus, this work constructed the CREER dataset, a large Corpus for Relation Extraction and Entity Recognition.

At first, we only wanted to get entity mentions and relations, so only these two annotations were kept. But during execution, we decided to keep all annotations for future research since the Stanford

Table 1: Comparison of existing corpora which contain entity and relation annotations.

| Dataset | Domain | #Sent. | Task | #Entity | #Relations |
|---|---|---|---|---|---|
| Wet Lab Protocols (Kulkarni et al., 2018) | biology | 14,301 | NER<br>RE | 60,745<br>60,745 | -<br>43,773 |
| CoNLL-2003 (Sang and De Meulder, 2003) | news | 20,744 | NER | 35,089 | - |
| SemEval-2010 Task8 (Hendrickx et al., 2019) | misc. | 10,717 | RE | 21,437 | 10,717 |
| OntoNotes 5.0 (Devlin et al., 2018;Weischedel et al., 2012) | misc. | 94,268 | Coref.<br>SRL<br>POS<br>Dep.<br>Consti. | 194,477<br>745,796<br>1,631,995<br>1,722,571<br>1,320,702 | 1,166,513<br>543,534<br>-<br>1,628,588<br>- |
| Penn Treebank (Marcus et al., 1994) | speech and news | 49,208<br>43,948<br>43,948 | POS<br>Dep.<br>Consti. | 1,173,766<br>1,090,777<br>871,264 | -<br>1,046,829<br>- |
| OIE2016 (Stanovsky and Dagan, 2016) | news and wiki | 2,534 | OpenIE | 15,717 | 12,451 |
| MPQA 3.0 (Deng and Wiebe, 2015) | news | 3,585 | ORL | 13,941 | 9,286 |
| SemEval-2014 Task4 (Maria Pontiki, 2014) | reviews | 4,451 | ABSA | 7,674 | - |
| **CREER** | **Wiki** | **144,732,654** | NER<br>RE | **371,186,870** | <br>**60,503,288** |

CoreNLP Annotator is a pipeline framework. Thus, this dataset provides all entity mentions and relation KBP in a sentence, as well as tokens, parse trees, and dependency tokens for parts of the Wikipedia corpus.

## 3 Data Collection

We follow the pre-training corpus used in BERT and GPT, and the CREER dataset is based on the English Wikipedia Dump dataset (2,500M words). Wikipedia is an online encyclopedia, and the online knowledge repository is edited by over tens of thousands of anonymous contributors. The Wikipedia documents are a good resource for fueling data science and natural language processing. We downloaded these documents from the Wikipedia dump (https://dumps.wikimedia.org/), and we used the text content of the English Wikipedia Dump dataset and ignored lists, tables, URLs, etc. in the dataset for the creation of the CREER dataset.

To obtain knowledge annotations, the work used off-the-shelf Stanford CoreNLP toolkits (Manning et al., 2014), which provide a stable and high-quality natural language analysis component, to annotate sentences in the Wikipedia dataset. The Stanford CoreNLP annotator provides Java API, and the lightweight framework is easy to use and customized to obtain the language annotation from plain text. This work used Library stanfordcorenlp, a Python wrapper for Stanford CoreNLP, to annotate linguistic annotation on the Wikipedia dataset. Although many large frameworks can be used to annotate multiple knowledge types from a text sentence, most of these toolkits require extra effort to get started with or do not meet my need.

## 4 Data Properties

Since the Stanford CoreNLP is a pipeline framework, several property results were produced before obtaining the relation Knowledge Base Population (KBP) and the entity mentions. At first, we followed the original motivation and preserved only entity mention and relation annotations. After the Stanford annotator had processed part of the text corpus, we decided to keep all knowledge annotations before obtaining the relation KBP and the entity mentions. Thus, this dataset contains complete annotated entity mentions and relation KBP and remains part of tokens, parse trees, and dependency tokens, which are annotated on the Wikipedia corpus. Appendix lists annotated examples of the CREER dataset, and several annotated properties in the CREER dataset are described below:

**tokenize** A token is an instance of a sentence, and an ordered set of tokens represents its semantics during language processing. Tokenization is the first step in the Stanford CoreNLP pipeline. The annotator broke a sentence or text into a sequence of tokens and preserved the character offsets for each token.

**ner** An entity is composed of a word or series of words and represents a semantic category. Stanford CoreNLP annotated named entities including PERSON, LOCATION, ORGANIZATION, and MISC using a conditional random fields model. Numerical entities such as MONEY and NUMBER were recognized using rule-based systems; the Stanford annotator also annotated SET and temporal properties including DATE, TIME, and DURATION. Figure 1 shows an example of a ner annotator.

**basic dependencies** A dependency represents a grammatical relationship for a sentence or text, and a unit of dependency is a triple which represents a relationship between a pair of word tokens. Figure 2 shows an example of a basic dependencies annotator.

**parse** A parse tree is a hierarchical structure used to represent the grammar of a sentence or text. The syntactic analysis includes constituent components and dependency components. Figure 3 shows an example of a parse annotator.

**knowledge base population (KBP)** A relation is defined by a pair of entities in a sentence. Stanford CoreNLP handles (1) slot filling and (2) entity linking to annotate a triplet of a relation and entities. Figure 4 shows an example of a relation KBP annotator.

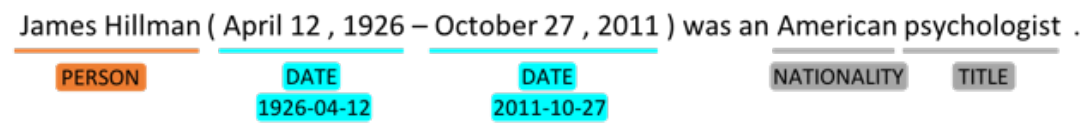

Figure 1: Visualization of an example result from ner annotator.

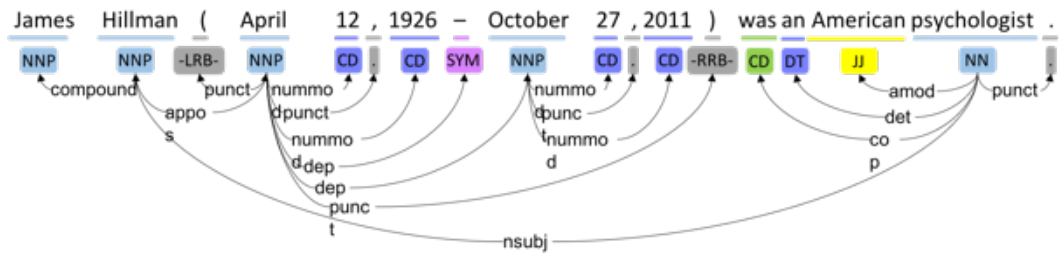

Figure 2: Visualization of an example result from basic dependencies annotator.

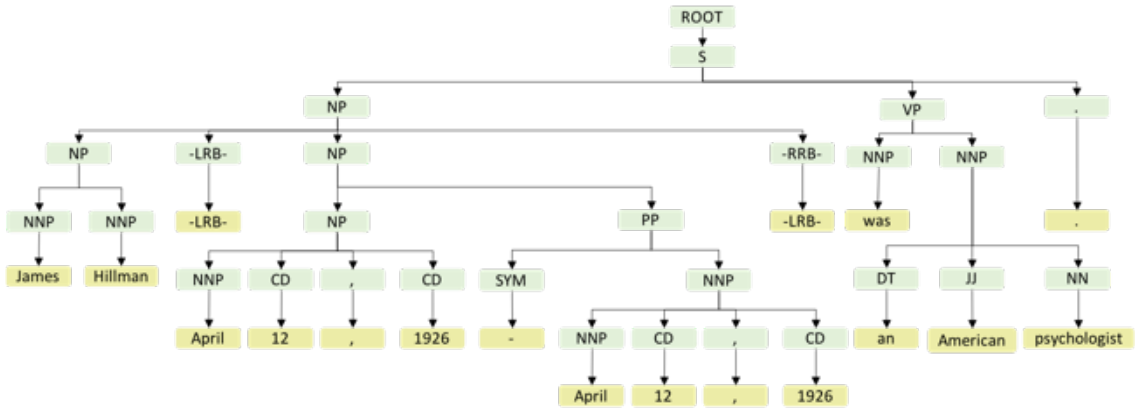

Figure 3: Visualization of an example result from parse annotator.

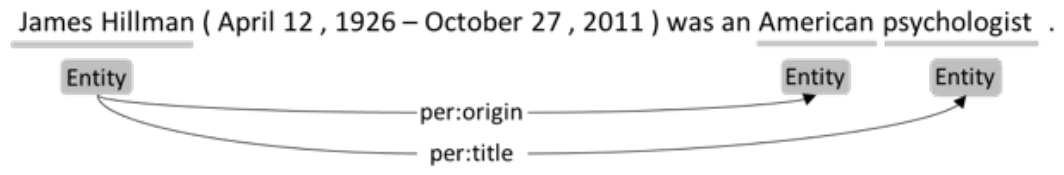

Figure 4: Visualization of an example result from relation KBP annotator.

This chapter does not describe all properties in the CREER dataset, and enhanced dependencies and enhanced++ dependencies have a higher-level semantic view than basic dependencies. See the documentation of Stanford CoreNLP for more details.

## 5 Conclusion

This paper describes the design of the CREER dataset, its initial motivation, data collection, and annotation. This dataset contains grammatical structures and world knowledge labels. This dataset follows widely used linguistic and semantic annotations, allowing for easy expansion of the dataset. The CREER dataset contains multiple linguistic and semantic annotations. The creation of the CREER dataset breaks the limitation that research can only inject knowledge into NLP systems in the form of unstructured knowledge (such as knowledge bases). This large supervised dataset can serve as a basic benchmark for future NLP systems and evaluations.

## Acknowledgments

We would like to thank the National Center for High-performance Computing (NCHC) of National Applied Research Laboratories (NARLabs) in Taiwan for providing computational and storage resources and the Ministry of Science and Technology of Taiwan for financially supporting this research with the project number of 108-2221-E-006 -103 -MY3.

## A Appendices

Since Stanford CoreNLP is a pipeline framework, there are additional annotation results before obtaining entity mentions and relation KBP.

| **Tokens** |
|---|
| **Sentence:**<br>The player can compete against computer opponents or, at least on the Atari ST and Amiga versions, with another player using two computers connected via a null modem cable, each with their own TV or monitor. |
| **Tokens**:<br>index: 4,<br>word: compete,<br>characterOffsetBegin: 15, characterOffsetEnd: 22,<br>Pos: VB,<br>ner: O<br>(The sentence is divided into several tokens, and we take one token as an example.) |

Table 2: Annotated examples of tokenizer annotator.

| **Entity Mentions** |
|---|
| **Sentence**:<br>The player can compete against computer opponents or, at least on the Atari ST and Amiga versions, with another player using two computers connected via a null modem cable, each with their own TV or monitor. |
| **Entity Mentions**:<br>docTokenBegin: 13<br>docTokenEnd: 15<br>tokenBegin: 13<br>tokenEnd: 15<br>text: Atari ST<br>characterOffsetBegin: 70<br>characterOffsetEnd: 78<br>ner: ORGANIZATION<br>(The annotator recognizes entities consisting of consecutive words, and we take an entity mention as an example.) |

Table 3: Annotated examples of ner annotator.

| **Relation (kbp)** |
|---|
| **Sentence**:<br>James Hillman (April 12, 1926 – October 27, 2011) was an American psychologist. |
| **Relation (kbp)**:<br>subject: James Hillman subjectSpan: [0, 2]<br>relation: per:title<br>relationSpan: [-2, -1]<br>object: psychologist objectSpan: [16, 17]<br>(The annotator extracts relation triplets meeting the TAC-KBP standard in a sentence, and we take one relation triplet as an example.) |

Table 4: Annotated examples of relation annotator.

**Parse Tree**

**Sentence**:
The player can compete against computer opponents or, at least on the Atari ST and Amiga versions, with another player using two computers connected via a null modem cable, each with their own TV or monitor.

**Parse Tree:**

```
(ROOT
 (S
  (NP (DT The) (NN player))
  (VP (MD can)
   (VP (VB compete)
    (PP
     (PP (IN against)
      (NP (NN computer) (NNS opponents)))
     (CC or)
     (, ,)
     (ADVP (IN at) (RBS least))
     (PP (IN on)
      (NP
       (NP (DT the) (NNP Atari) (NNP ST))
       (CC and)
       (NP (NNP Amiga) (NNS versions)))))
    (, ,)
    (PP (IN with)
     (NP
      (NP (DT another) (NN player))
      (VP (VBG using)
       (NP
        (NP (CD two) (NNS computers))
        (VP (VBN connected)
         (PP (IN via)
          (NP
           (NP (DT a) (JJ null) (NN modem) (NN cable))
           (, ,)
           (NP
            (NP (DT each))
            (PP (IN with)
             (NP (PRP$ their) (JJ own) (NN TV)
              (CC or)
              (NN monitor))))))))))))
                 …
```

(The annotator constructs a syntax tree from a sentence, and we visualize part of the parse tree.)

Table 5: Annotated examples of parse annotator.